\documentclass[conference,a4paper,finalsubmission]{IEEEtran}

\usepackage{color}
\usepackage{cite}
\usepackage{comment}
\usepackage[a4paper, top=122pt, left=54pt, right=33pt, bottom=54pt]{geometry}
\usepackage{amsmath, amssymb}
\usepackage{algorithm}
\usepackage{graphicx}
\usepackage{fancyhdr}
\usepackage{lipsum}
\usepackage[noend]{algpseudocode}

\IEEEoverridecommandlockouts                              

\title{
Integrating Robotic Navigation with Blockchain: A Novel PoS-Based Approach for Heterogeneous Robotic Teams
}

\author{Nasim~Paykari, Ali~Alfatemi, Damian M. Lyons,~\IEEEmembership{Senior~Member,~IEEE,}
        Mohamed~Rahouti,~\IEEEmembership{Member,~IEEE}
\thanks{N. Paykari, A. Alfatemi, D. M. Lyons, and M. Rahouti are with the department of Computer \& Information Science, Fordham University, NY, USA. e-mail: \{npaykari, aalfatemi, dlyons, mrahouti\}@fordham.edu.\newline This work was presented at 2024 21st International Conference on Ubiquitous Robots (UR) June 24 - 27, 2024, New York University, USA}
}

\fancypagestyle{firstpage}{
  \fancyhf{} 
  \fancyhead[C]{\textbf{2024 21st International Conference on Ubiquitous Robots (UR) June 24 - 27, 2024, New York University, USA}} 
  \fancyfoot[C]{\textbf{979-8-3503-6107-0/24/\$31.00 ©2024 IEEE}} 
}

\begin{document}

\maketitle
\thispagestyle{empty}
\pagestyle{empty}

\begin{abstract}

This work explores a novel integration of blockchain methodologies with Wide Area Visual Navigation (WAVN) to address challenges in visual navigation for a heterogeneous team of mobile robots deployed for unstructured applications in agriculture, forestry, etc. Focusing on overcoming challenges such as GPS independence, environmental changes, and computational limitations, the study introduces the Proof of Stake (PoS) mechanism, commonly used in blockchain systems, into the WAVN framework \cite{Lyons_2022}. This integration aims to enhance the cooperative navigation capabilities of robotic teams by prioritizing robot contributions based on their navigation reliability. The methodology involves a stake weight function, consensus score with PoS, and a navigability function, addressing the computational complexities of robotic cooperation and data validation. This innovative approach promises to optimize robotic teamwork by leveraging blockchain principles, offering insights into the scalability, efficiency, and overall system performance. The project anticipates significant advancements in autonomous navigation and the broader application of blockchain technology beyond its traditional financial context.

\end{abstract}

\section{Introduction}

In the realm of autonomous robotics, the need for efficient and reliable navigation is paramount. Traditional navigation systems heavily depend on GPS and other sensor-based technologies. While these systems perform well in environments with stable and predictable conditions, they face significant challenges in environments where GPS is unreliable or unavailable or areas with dynamic and unpredictable changes, such as unstructured agricultural or forestry applications. Moreover, the navigation of heterogeneous robotic teams adds computational complexity due to different robots' varying capabilities and sensor setups. Although beneficial for task versatility, this diversity challenges cooperative navigation and data integration. The state-of-the-art solutions have explored sensor fusion and machine learning techniques to enhance navigation capabilities. However, these methods often encounter computational efficiency and scalability limitations, mainly when dealing with large-scale, diverse robotic teams.

The motivation for this project arises from the need to address these challenges with innovative solutions. Integrating blockchain methodologies into Wide Area Visual Navigation (WAVN) systems for robotic teams presents a unique solution. Blockchain technology, known for its secure, transparent, and decentralized nature in financial applications, offers untapped potential in robotic navigation as well as other area \cite{jahannia2024optical, miah2023blockchain}. By leveraging blockchain's trust and verification mechanisms, particularly the Proof of Stake (PoS) concept, this project aims to enhance the reliability and efficiency of data used for navigation decisions. PoS allows for prioritizing data from more reliable navigation sources within the robotic team, thereby improving the overall navigation accuracy and reducing the reliance on traditional, centralized data verification methods. This integration not only promises to tackle the computational and scalability challenges but also introduces a novel approach to managing and validating navigation data in real time, fostering cooperative decision-making among diverse robotic entities.

The project is driven by the broader goal of advancing the field of autonomous robotics, specifically in complex and dynamic environments. By addressing the current limitations and exploring the synergy between blockchain technologies and robotic navigation systems, the project aims to set new benchmarks in autonomous navigation systems' efficiency, reliability, and scalability, potentially revolutionizing how robotic teams operate in various industrial and service applications.

\section{State-of-the-art}

The state-of-the-art autonomous navigation field for heterogeneous robotic teams, particularly in applications such as precision agriculture, has been rapidly evolving \cite{st2020design}. Current research primarily focuses on enhancing navigation capabilities under challenging conditions, like GPS denial and dynamic environmental changes \cite{azpurua2023survey}. Traditional approaches rely heavily on GPS and other sensor-based navigation systems, which, although effective in stable conditions, struggle in environments with limited or no GPS availability and fluctuating terrain. Recent advancements have introduced more adaptive and robust methods, such as machine learning algorithms and sensor fusion techniques, which integrate data from various sources (like LIDAR, cameras, and inertial sensors) to improve navigation accuracy and reliability \cite{Druen_2020}. These methods, however, often face limitations in terms of computational efficiency and scalability, especially when deployed in large-scale environments with multiple robots.

\newgeometry{a4paper, top=104pt, left=54pt, right=33pt, bottom=54pt}
In response to these challenges, integrating blockchain methodologies into WAVN systems presents a groundbreaking direction \cite{rahouti2022lightweight, rahouti2023decentralized}. Blockchain technology, traditionally associated with secure and decentralized financial transactions \cite{ali2019blockchain}, offers unique advantages in robotic navigation, particularly regarding data validation and trustworthiness \cite{aditya2021survey, paykari2023assessing}. Introducing the PoS mechanism into the WAVN framework is a novel approach, prioritizing robot data contributions based on their navigation reliability \cite{krishnamohan2022analysing}. This methodology promotes a decentralized yet reliable consensus mechanism, ensuring that the most accurate and reliable navigation data is used for decision-making processes. The application of blockchain in this context is not just about secure data handling; it's about enhancing cooperative decision-making and efficiency in robotic teams \cite{lyons2023wavn}. This approach also addresses significant computational challenges, as the PoS mechanism can potentially reduce the computational overhead compared to traditional blockchain consensus mechanisms like Proof of Work (PoW). As such, this integration addresses the current limitations in autonomous navigation systems and opens new avenues for applying blockchain technology beyond its conventional financial applications.

Unlike existing solutions, this paper presents a work-in-progress project on integrating blockchain technology, with a unique/novel PoS consensus, into WAVN to enhance the cooperative navigation of mobile robots, especially for precision agriculture. This effort aims to address challenges like GPS denial and environmental changes. It aims to optimize robotic teamwork and autonomous navigation by applying a lightweight, WAVN-specialized consensus for improved scalability, efficiency, and system performance.

\section{Methodology}

The methodology presented in this work-in-progress project involves integrating blockchain technology, specifically PoS, with WAVN to enhance the navigation of a heterogeneous team of mobile robots. This integration focuses on leveraging blockchain's trust and verification mechanisms to prioritize reliable navigation data. The methodology includes a stake weight function, a consensus score with PoS, and a navigability function, addressing the challenges of cooperative navigation and real-time data validation. This innovative approach aims to optimize robotic teamwork, scalability, efficiency, and overall system performance in environments like precision agriculture. The proposed methodology is outlined next.

Given a set of robots \( R = \{ r_1, r_2, ..., r_n \} \), and their respective stakes \( S = \{ s_1, s_2, ..., s_n \} \) where \( s_i \) represents the stake of robot \( r_i \) (interpreted as its navigation reliability):

\begin{enumerate}
    \item \textbf{Stake weight function}:
    \begin{equation}
      W(r_i) = \frac{s_i}{\sum_{j=1}^{n} s_j}  
    \end{equation}
    This function calculates the weight of each robot based on its stake relative to the total stake in the system.
    
    \item \textbf{Consensus score with PoS}:
    \begin{equation} \label{eq:Consensus Score with PoS}
      C_{PoS}(r_i, r_j) = W(r_i) \times \sum_{k=1}^{m} \omega_k \times I(l_k, r_i, r_j)  
    \end{equation}
    Where \( \omega_k \) is the weight of the landmark \( l_k \) and \( I \) is the indicator function defined as:

        \[ I(l_k, r_i, r_j) = 
    \begin{cases} 
        1 & \text{if both } r_i \text{ and } r_j \text{ recognize } l_k \\
        0 & \text{otherwise} 
    \end{cases} \]
    
    \item \textbf{Navigability function with PoS}:
    \begin{equation} \label{eq:Navigability Function with PoS}
     N_{PoS}(r_i) = \sum_{j=1, j\neq i}^{n} \alpha_{ij} \times C_{PoS}(r_i, r_j) 
    \end{equation}
    Where \( \alpha_{ij} \) represents the relative importance or trustworthiness of robot \( r_j \) to \( r_i \) as defined previously.
\end{enumerate}

The consensus score function \( C(r_i, r_j) \) iterates over all landmarks \( m \) for each pair of robots. For \( n \) robots and \( m \) landmarks, the complexity is \( O(m  \binom{n}{2}) \).

The navigability function \( N(r_i) \) for a single robot involves calculating the consensus score with all other \( n-1 \) robots, leading to a complexity of \( O(n m (n-1)) \), which approximates to \( O(n^2 m) \).

In the context of blockchain integration with robotic navigation, the project outlines two main areas of computational complexity: the consensus mechanism and data logging.
\begin{itemize}
    \item Consensus mechanism: The computational complexity here is twofold, depending on both the consensus score function and the frequency of blockchain updates. The consensus score function, integral to the decision-making process, increases in complexity with the number of robots and landmarks involved. This function influences how robots reach a collective decision based on their individual stakes and the environmental data they gather. The frequency of blockchain updates adds another layer of computational complexity. It is important to note that more frequent updates can provide more current data, enhancing navigation accuracy, increasing computational demands, and potentially slowing down the system.
    \item Data logging: This aspect's computational complexity is proportional to the number of robots and the frequency of updates. Each robot generates data that needs to be logged and verified through the blockchain, which can be an intensive process, especially with a large number of robots. While beneficial for real-time navigation accuracy, frequent updates can exacerbate this computational complexity by increasing the volume of data that needs to be processed and recorded on the blockchain. This challenge highlights a critical balance between ensuring timely and accurate navigation data and effectively managing the computational and storage resources.
\end{itemize}

Furthermore, optimizing the WAVN algorithm to leverage the blockchain ledger involves integrating the consensus and navigability functions. The computational complexity is tied to these integrations and the efficiency of the WAVN algorithm.

Last, the overall computational complexity combines the computational complexities of the above components, dominated by the consensus score and navigability functions, with complexities of \( O(m \binom{n}{2}) \) and \( O(n^2 m) \), respectively.

\section{Preliminary Results}
\begin{figure}[h]
    \centering
    \includegraphics[width=0.40\textwidth]{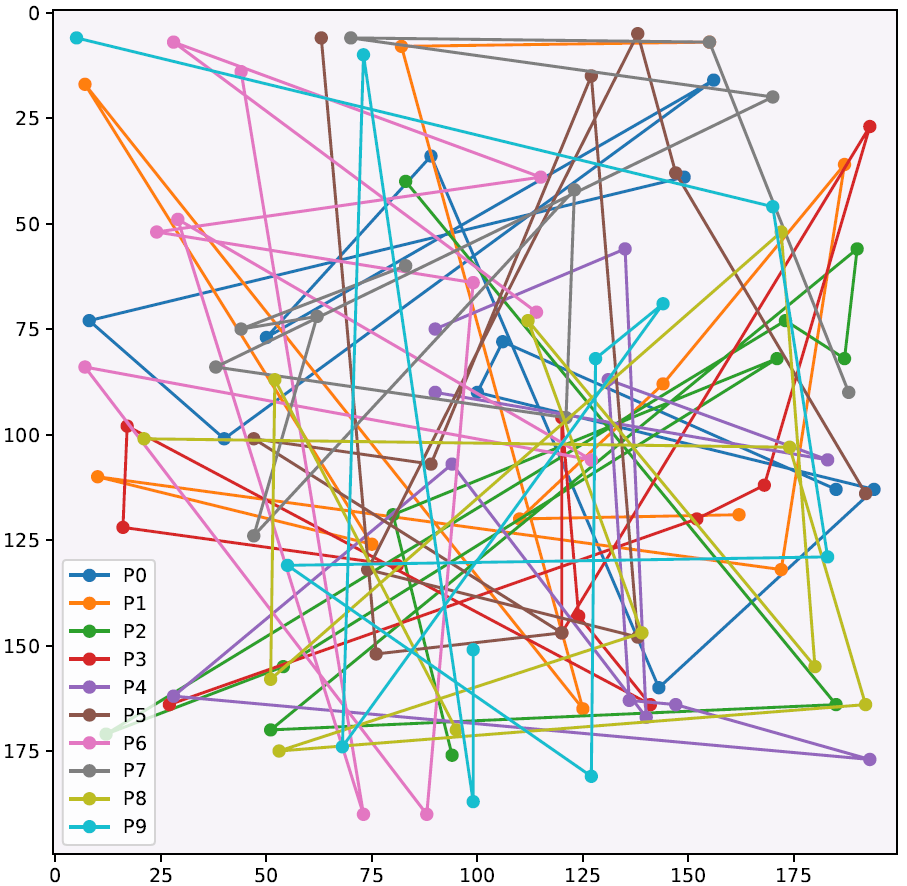}
    \caption{The trajectory of each robot throughout the experiment illustrates how the position of each robot has evolved.}
    \label{fig:movements}
\end{figure}

In this section, a preliminary experiment is conducted to demonstrate how the navigability factor is influenced by the interaction within a team of robots and how it is affected by the quality of landmarks. The experiment involved 10 robots and 20 landmarks, randomly placed in a world with dimensions of $200\times{}200$. In the subsequent step, within 10 loops, the location of robots changes randomly, as depicted in Figure \ref{fig:movements}. Randomness is employed throughout to simulate the process, ensuring maximum stress on the experiment when a parameter is absent.

\begin{figure}[h]
    \centering
    \includegraphics[width=0.35\textwidth]{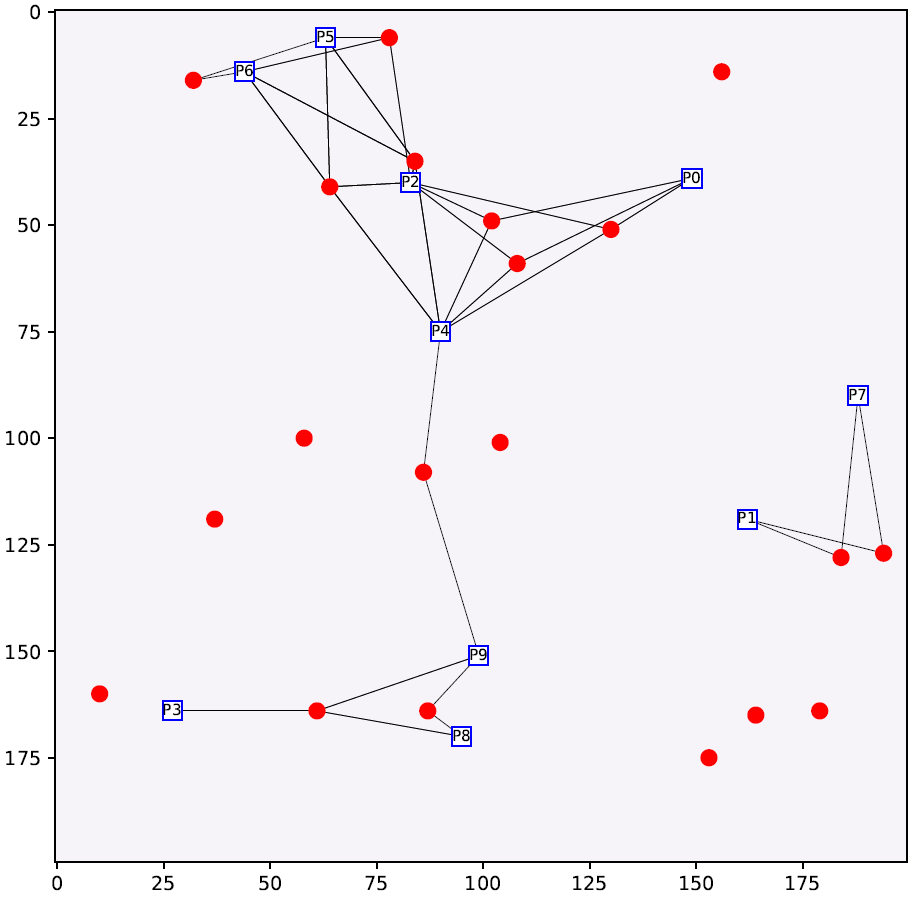}
    \includegraphics[width=0.35\textwidth]{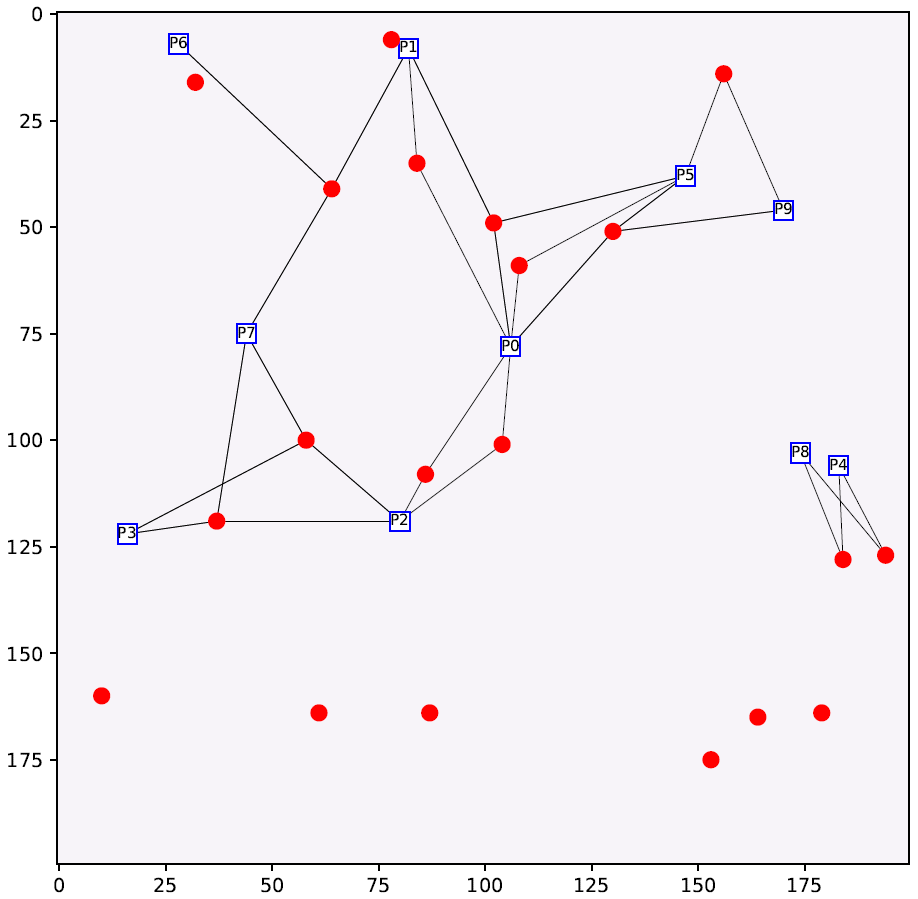}
    \caption{Arrangement of robots in the initial (above) and final loops (below). Red circles represent the positions of landmarks, while squares denote robots that move randomly. Lines connect common landmarks to the corresponding robots, illustrating the relationships between each pair of robots and their common landmarks.}
    \label{fig:loops}
\end{figure}

The results indicate the generation of 45 blocks and 458 transactions. The maximum number of common landmarks between two robots is 5, while the minimum is zero. Figure \ref{fig:loops} displays the graphs illustrating the positions of robots and common landmarks pairwise in the first and last loops. Additionally, the outcomes reveal that robots have been selected as generators of blocks, ranging from a maximum of 9 for P4 and P5 to a minimum of 1 for P3. Given the absence of an actual panorama or point of view, we randomly assign the quality of common landmarks between a pair of robots, which serves as \(w_k\) of navigability as defined in Equation \ref{eq:Consensus Score with PoS}.

\begin{figure}[h]
    \centering
    \includegraphics[width=0.45\textwidth]{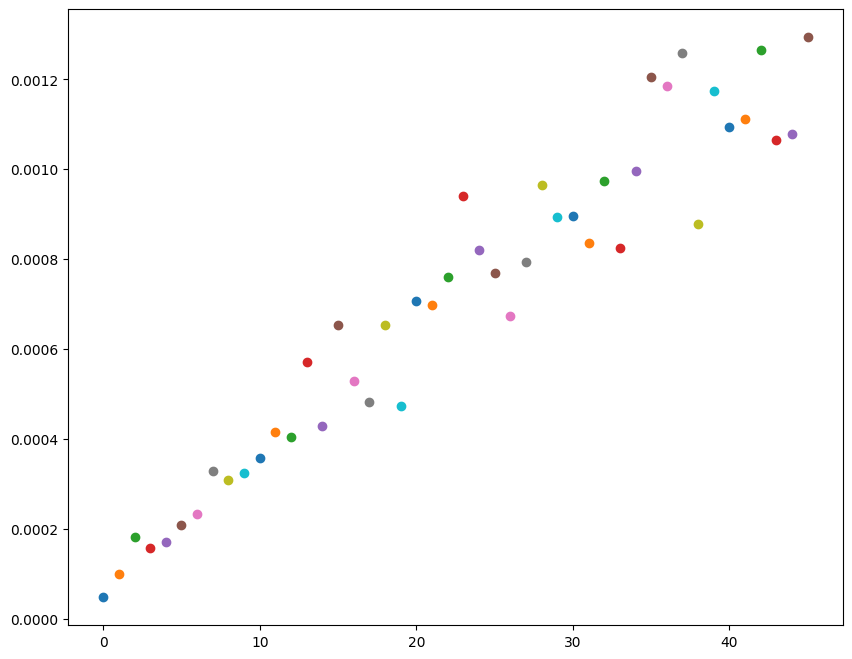}
    \caption{Navigability over time (blocks). Colors represent the different blocks. Over time, the number of blocks grows, and the navigability matrix values within each block experience changes. Each point on the graph represents the average values of elements in the navigability matrix at the block's generation time. The intervals between these points are unequal and align with the generation of blocks over time.}
    \label{fig:navi}
\end{figure}

Navigability is the final parameter affecting the block miner and the optimal robot for pathfinding. We incorporate the significance of robots, denoted as $\alpha$ in Equation \ref{eq:Navigability Function with PoS}, by examining past transactions to determine the cooperation of each robot with the main robot searching for a home. The higher the number of transactions between them, the more important the robot is considered. To prevent an increase in search time with a large team of robots, we limit our analysis to the top 10 transactions and assign them the maximum level of importance. Consequently, we have 10 levels of importance for each robot concerning the robot searching for a home. The graph given in Figure \ref{fig:navi} illustrates how the average navigability between robots increases over time.

It's worth noting while the navigability between pairs increases over time, a decrease in the quality of matches of common landmarks between two robots has a noticeable impact, leading to a valley in the navigability graph. An interesting observation is that navigability grows when the quality of common landmarks between these two robots is restored.

\section{Discussion}

The work-in-progress on integrating blockchain with robotic navigation faces distinct challenges. A primary challenge involves the dynamics of stake management, explicitly determining the appropriate methods and timing for adjusting the stakes of individual robots to reflect their navigation reliability accurately. Scalability is another concern, as the system's efficiency could be impacted by the increasing number of robots (n) and landmarks (m), potentially straining the computational resources. Furthermore, optimizing the consensus and navigability functions for maximum efficiency presents a significant challenge, as these are core to the system's decision-making process. The integration of blockchain also introduces additional overhead, which could affect the system's overall performance. Addressing these challenges is crucial to ensure the practicality and effectiveness of this innovative approach.

Regarding limitations, the project confronts several inherent constraints. The computational complexity of the consensus score function and the navigability function, which form the system's backbone, escalates significantly as the number of robots and landmarks increases. This increase in computational complexity could lead to efficiency issues, particularly in large-scale environments. Additionally, integrating blockchain technology while offering advantages in trust and verification introduces additional computational and communication overhead. This overhead is especially pertinent in terms of blockchain update frequencies and the data logging requirements associated with a large number of robots. The reliance on the PoS mechanism also necessitates careful calibration and management of stakes, which could be complex and time-consuming. These limitations underline the need for further methodology refinement to ensure scalability, efficiency, and real-world applicability.

\section{Conclusion and Future Work}

Integrating the PoS mechanism into robotic navigation systems presents a novel approach to enhancing cooperative navigation. This research will provide valuable insights into the potential applications of blockchain principles outside their traditional domains.

The future directions of this project, integrating blockchain methodologies with Wide Area Visual Navigation (WAVN) in heterogeneous robotic teams, are poised to open new frontiers in autonomous systems. A key focus will be on refining and scaling the blockchain-WAVN integration to handle larger and more complex robotic teams, adapting to a broader range of environments beyond precision agriculture. This includes exploring advanced blockchain algorithms that can further optimize the efficiency and reduce the computational overhead of the system. Additionally, the integration of AI and machine learning techniques with the PoS-based WAVN system offers a promising avenue to enhance decision-making algorithms, enabling more dynamic and adaptive navigation strategies. 

Another significant direction will be the development of robust security protocols within the blockchain framework to safeguard against potential cyber threats, a critical consideration as the system’s complexity increases. Finally, real-world testing and validation of the system in various industrial scenarios will be crucial to understand its practical limitations and opportunities, paving the way for its adaptation in other fields where autonomous navigation is essential, such as search and rescue operations, environmental monitoring, and urban planning. This project has the potential not only to advance the field of autonomous robotics but also to provide valuable insights into the versatile applications of blockchain technology.


\bibliographystyle{IEEEtran}
\bibliography{refs}



\end{document}